\documentclass[letterpaper]{article} % DO NOT CHANGE THIS
\usepackage{aaai25}  % DO NOT CHANGE THIS
\usepackage{times}  % DO NOT CHANGE THIS
\usepackage{helvet}  % DO NOT CHANGE THIS
\usepackage{courier}  % DO NOT CHANGE THIS
\usepackage[hyphens]{url}  % DO NOT CHANGE THIS
\usepackage{graphicx} % DO NOT CHANGE THIS
\usepackage{natbib}  % DO NOT CHANGE THIS AND DO NOT ADD ANY OPTIONS TO IT
\usepackage{caption} % DO NOT CHANGE THIS AND DO NOT ADD ANY OPTIONS TO IT
\usepackage{algorithm}
\usepackage{algorithmic}

\usepackage{multirow}
\usepackage{booktabs}
\usepackage{breqn}
\usepackage{flexisym}
\usepackage{mathtools}
\usepackage{newfloat}
\usepackage{listings}
\DeclareCaptionStyle{ruled}{labelfont=normalfont,labelsep=colon,strut=off} % DO NOT CHANGE THIS
\floatstyle{ruled}
\newfloat{listing}{tb}{lst}{}
\floatname{listing}{Listing}
\newtheorem{theorem}{Theorem}

\usepackage{xspace}
\newcommand{\ie}{\emph{i.e.,}\xspace}
\newcommand{\eg}{\emph{e.g.,}\xspace}

\title{
SoLA: Leveraging Soft Activation Sparsity and Low-Rank Decomposition for Large Language Model Compression
}
\author{
    Xinhao Huang \textsuperscript{\rm 1},
    You-Liang Huang \textsuperscript{\rm 1},
    Zeyi Wen\textsuperscript{\thanks{Corresponding Author}\rm 1, 2}
}
\affiliations{
    \textsuperscript{\rm 1}The Hong Kong University of Science and Technology (Guangzhou) \\ 
    \textsuperscript{\rm 2}The Hong Kong University of Science and Technology \\
    \{xhuang171, yhuang142\}@connect.hkust-gz.edu.cn, wenzeyi@ust.hk
}

\usepackage{bibentry}
\begin{document}

\maketitle

\begin{abstract}

Large language models (LLMs) have demonstrated impressive capabilities across various tasks, but the billion-scale parameters pose deployment challenges. 
Although existing methods attempt to reduce the scale of LLMs, 
they require either special hardware support or expensive post-training to maintain model quality. 
To facilitate efficient and affordable model slimming, we propose a novel training-free compression method for LLMs, named ``SoLA'', which leverages \textbf{So}ft activation sparsity and \textbf{L}ow-r\textbf{A}nk decomposition. SoLA can identify and retain a minority of components significantly contributing to inference, while compressing the majority through low-rank decomposition, based on our analysis of the activation pattern in the feed-forward network (FFN) of modern LLMs.
To alleviate the decomposition loss, SoLA is equipped with an adaptive component-wise low-rank allocation strategy to assign appropriate truncation positions for different weight matrices.
We conduct extensive experiments on LLaMA-2-7B/13B/70B and Mistral-7B models across a variety of benchmarks.
SoLA exhibits remarkable improvement in both language modeling and downstream task accuracy without post-training. For example,
with a 30\% compression rate on the LLaMA-2-70B model, SoLA surpasses the state-of-the-art method by reducing perplexity from 6.95 to 4.44 and enhancing downstream task accuracy by 10\%.

\end{abstract}

% Uncomment the following to link to your code, datasets, an extended version or similar.
\begin{links}
    \link{Code}{https://github.com/xinhaoH/SoLA}
\end{links}

\section{Introduction}

In recent years, the capabilities of large language models (LLMs) based on Transformers have been widely demonstrated across diverse tasks, and their sizes tend to continuously increase to improve performance according to the scaling law~\cite{scaling_laws}. These LLMs with a large number of parameters demand significant storage and computation resources, posing obstacles to their deployment and utilization. Researchers attempt to mitigate the cost of LLMs by reducing model parameters with compression methods. The predominant compression techniques include unstructured pruning, structured pruning, quantization, and low-rank decomposition.

Unstructured pruning exploits the inherent sparsity of the model to remove certain weights.
However, several concerns impede its usability, including unavailable activation sparsity due to modification of activation functions (\eg replace ReLU with SiLU) and the lack of hardware support on commodity devices~\cite{dejavu, wanda}.
In comparison, structured pruning removes entire channels or other structured components from LLMs, which leads to notable precision degradation because of aggressive modification to the model structure, requiring fine-tuning to recover performance~\citep{llm_pruner}.
Different from pruning, quantization aims to reduce memory consumption through storing model parameters in low-bit floating point numbers, which can be incorporated into fine-tuning for better accuracy recovery~\cite {gptq}. 

Compared with pruning and quantization methods, low-rank decomposition compression techniques, such as Singular Value Decomposition (SVD), do not need special hardware support or expensive retraining, by using lower-rank matrices to approximate the weight matrix in LLMs. 
However, the existing approach exhibits significant performance degradation due to high compression loss~\cite{asvd}. This reduction in performance is exacerbated by ignoring data distribution in inputs and outputs~\cite{svd_llm}, as well as missing the consideration for the differences among model components (\ie weight matrices of feed-forward and attention module).

In this work, we propose a novel training-free compression method for LLMs, namely SoLA, which leverages soft activation sparsity and low-rank decomposition.
SoLA first recognizes and retains a small part of neurons (\eg 15\%) with high activation norms in the FFN, which contributes to the majority of the model performance during inference. Then, SoLA applies low-rank decomposition to compress the weight matrices corresponding to the rest of the neurons. To further boost the model quality after compression, SoLA exploits an adaptive rank allocation strategy for assessing the decomposition quality and determining the truncation position for each type of weight matrix, since different types of weight matrices exhibit varying levels of sensitivity to compression~\cite{svd_llm}.

We compare SoLA with the state-of-the-art pruning and low-rank decomposition methods.
To demonstrate SoLA's generability, we conduct evaluations across a variety of benchmarks using different LLM families (LLaMA-2 and Mistral) at three scales (7B, 13B, and 70B).
The experimental results show that SoLA preserves the generation quality and achieves remarkable downstream task accuracy at different compression rate levels. 
For instance, in a 30\% compression ratio scenario with LLaMA-2-70B, SoLA outperforms existing state-of-the-art methods, achieving a perplexity reduction from 6.95 to 4.44 and a 10\% improvement in downstream task accuracy.

Our contributions can be summarized as follows:
\begin{itemize}
    \item We introduce SoLA, a training-free compression method utilizing soft activation sparsity and low-rank decomposition.
    We analyze the soft activation sparsity in the FFN of modern LLMs and achieve fine-grained compression.
    \item We propose an adaptive component-wise low-rank allocation strategy that considers the differences between weight matrices and allocates appropriate truncation positions for different types of weight matrices, achieving enhanced model quality after compression, even with high compression ratios.
    \item Extensive experiments show that SoLA achieves remarkable performance in perplexity and widely-used benchmarks, and outperforms the state-of-the-art method without post-training.
\end{itemize}

\section{Related Works}

In this section, we review related compression techniques, including network pruning, model quantization, and low-rank decomposition, as essential strategies to mitigate the burden imposed by large-scale models during inference.

\subsection{Network Pruning and Quantization Methods}

Network pruning includes non-structured pruning and structured pruning based on the paradigm of network parameter reduction.
Recent studies on unstructured pruning have concentrated on the sparsity of the LLM weight matrices, pruning the model by eliminating certain weights. 
Dejavu~\cite{dejavu} omits the computation of weight matrices corresponding to the ReLU zero activation value.
SparseGPT, proposed by~\citet{sparse_gpt}, decomposes the pruning problem to a set of extremely large-scale instances of sparse regression.
Wanda~\cite{wanda} computes weight importance metric utilizing weights and activations to induce sparsity in pretrained LLMs.
\citet{pruner_zero} employ genetic programming to identify optimized symbolic pruning metrics suitable for LLMs.
However, the current mainstream models no longer employ ReLU, and thus cannot leverage the sparsity of zero activations. 
Moreover, the present hardware ecosystem does not adequately support unstructured pruning~\cite{wanda}.

In structured pruning methods,
\citet{llm_pruner} evaluate the importance of each structure through a first-order Taylor expansion and prunes the structures with the lowest scores.
LLM Surgeon~\cite{llm_surgeon} achieves pruning of LLMs by extending the second-order Hessian approximation method of the Kronecker factorized Fisher information matrix.
FLAP~\cite{flap} designs a fluctuation pruning metric and then introduces a bias term to recover the output feature map.
\citet{slice_gpt} utilize a transformation matrix Q to remove rows and columns of the weight matrix but requires additional adapters to handle the reduced dimensions.
Some methods~\cite{short_gpt, delete_layers} directly remove layers in the model that have similar inputs and outputs, but this can result in significant performance degradation, especially when the prune ratio exceeds 20\%. 

Quantization methods achieve memory consumption reduction through storing model parameters in low-bit floating point numbers.
Gptq~\cite{gptq} uses inverse Hessian information to weight quantization.
Qlora, as presented by~\cite{qlora}, fine-tunes low-rank adapters by backpropagating gradients through a frozen 4-bit quantized network.
But for better accuracy recovery, quantization techniques tend to need a subsequent fine-tuning process.

\subsection{Low-Rank Decomposition}
% Approximation
In the low-rank decomposition approach, the weight matrix is replaced by the product of two smaller matrices.
One category of methods decomposes the weight matrix using SVD or its variants.
\citet{fwsvd} utilizes Fisher information to measure the importance of parameters, but the high computational cost is incurred due to gradient computation.
ASVD~\cite{asvd} uses a diagonal matrix to represent the influence of input channels on weights, eliminating the need for gradient computation. 
SVD-LLM~\cite{svd_llm} establishes a direct relationship between singular values and compression loss, choosing the truncation of singular values with minimal compression loss.
\citet{ffsplit} notice the imbalance of activation norms in BERT, and leverage this feature in model decomposition. However, it ignores module differences and data distribution in inputs and outputs, which could cause drastic performance degradation in modern LLMs~\citep{asvd}.  %lead to

Another category of methods performs decomposition in the feature space.
\citet{features_low_rank} propose Atomic Feature Mimicking (AFM), which uses PCA decomposition to decompose the output vector (\ie the product of weights and inputs of the fully connected layer).
LORD also employs AFM for low-rank decomposition, which is applied in monolingual code generation.
Building upon~\citet{features_low_rank}, Bolaco~\cite{bolaco} utilizes Bayesian optimization to search for an appropriate truncation position.
To attain optimal performance, these feature-based methods need to precisely estimate feature distribution in extremely high dimensional feature space, which is difficult for tens of billions of scale LLMs.

\begin{figure*}[t]
\centering
\includegraphics[width=0.95\linewidth]{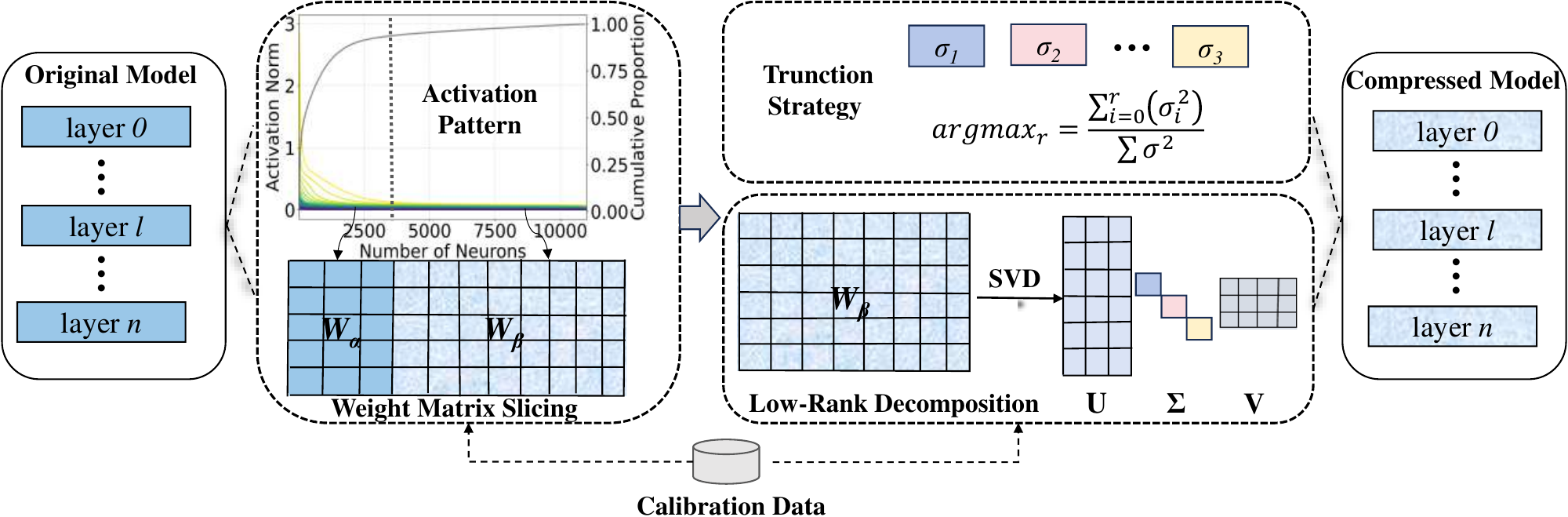} 
\caption{Framework of the proposed SoLA. We initially recognize the soft activation sparsity within the feed-forward network. Leveraging this property, we introduce a fine-grained model decomposition technique to preserve model quality. Furthermore, to alleviate the compression error of SVD, we develop an adaptive component-wise truncation strategy to allocate appropriate truncation positions for different types of weight matrices.}
\label{fig:overview}
\end{figure*}

\section{Preliminaries}

In this section, we briefly explain the computation process of the feed-forward network and then present the concept of `neuron' used in this paper. In the end, we introduce the foundation of low-rank decomposition.

\textbf{Feed-Forward Network}: 
To facilitate subsequent demonstrations,
we formalize the computation process of a two-layer feed-forward network (FFN) in Transformers.
Given the hidden dimension $d$ and the intermediate dimension $d_{ff}$, the sequential computation of two linear layers FFN can be formalized as:
\begin{equation}
  h = \sigma (X W^{in})
  \label{eq1}
\end{equation}
\begin{equation}
  out = h W^{out}
  \label{eq2}
\end{equation}
where $X \in R^{d}$ represents the input, $\sigma$ denotes the activation function, \eg SiLU and GeLU. The intermediate state is denoted by $h \in R^{d_{ff}}$, $out \in R^{d}$, and the weight matrices are defined as $W^{in} \in R^{d \times d_{ff}}$ and $W^{out} \in R^{d_{ff} \times d}$.
We omit bias terms for convenience.

\textbf{Neuron}:
In the context of the FFN, the term `neuron' denotes an element of the intermediate state.
Specifically, the $i$-th neuron corresponds to the $i$-th element of the intermediate state $h$. For a given weight matrix $W$, the notation $W_{i,:}$ denotes the $i$-th row, representing the weights leading to the $i$-th neuron, while $W_{:,i}$ indicates the $i$-th column, representing the weights emanating from the $i$-th neuron. 
In Equations~\eqref{eq1} and~\eqref{eq2}, the $i$-th column of the input weight matrix $W^{in}$ and the $i$-th row of the output weight matrix $W^{out}$ corresponding to the $i$-th neuron.

\textbf{Low-Rank Decomposition}: 
Given a weight matrix $W \in R^{m \times n}$, we can apply Singular Value Decomposition (SVD) to decompose $W$ into:
\begin{equation}
  W = U \Sigma V
  \label{eq3}
\end{equation}
where $U \in R^{m \times m}$, $V \in R^{n \times n}$, and $\Sigma \in R^{m \times n}$ is a rectangular diagonal matrix whose diagonal elements are singular values arranged in descending order. 

The matrix $W$ can be approximated by the largest $k$ singular values ($k < n$), and then:
\begin{equation}
  W \approx AB 
  \label{eq4}
\end{equation}
where $A = (U_k \sqrt{\Sigma_k})$, $B = (\sqrt{\Sigma_k} V_k^T)$, $U_k \in R^{m \times k}$ and $V_k^T \in R^{k \times n}$ are the rank-$k$ approximation matrices, and $\sqrt{\Sigma_k} \in R^{k \times k}$ is a diagonal matrix by the square-roots of the corresponding top-$k$ singular values in $\Sigma$.

When employing SVD to decompose the weight matrix of LLMs into approximate matrices, opting for a smaller value of $k$ results in a significant accuracy drop, whereas a larger $k$ increases the model size. The reconstruction loss can be formulated as follows:
\begin{equation}
  L = \Vert W - W' \Vert _F
  \label{eq6}
\end{equation}
where Equation~\eqref{eq4} can be applied to $W'$ to approximate $W$. 
This low-rank approximation reduces the number of parameters from $m \times n$ to $(m + n) \times k$.

\section{Methodology}

As shown in Figure \ref{fig:overview}, we first recognize and analyze patterns of activation norms in the FFN of modern LLMs. Then, based on the analysis and the properties, we introduce a fine-grained model decomposition method that leverages both activation awareness and soft activation sparsity to retain the model quality. To further mitigate reconstruction error brought by model decomposition, we devise an adaptive component-wise low-rank allocation strategy to determine the desired truncation position of each component.

\begin{figure}[t]
\centering
\includegraphics[width=0.99\columnwidth]{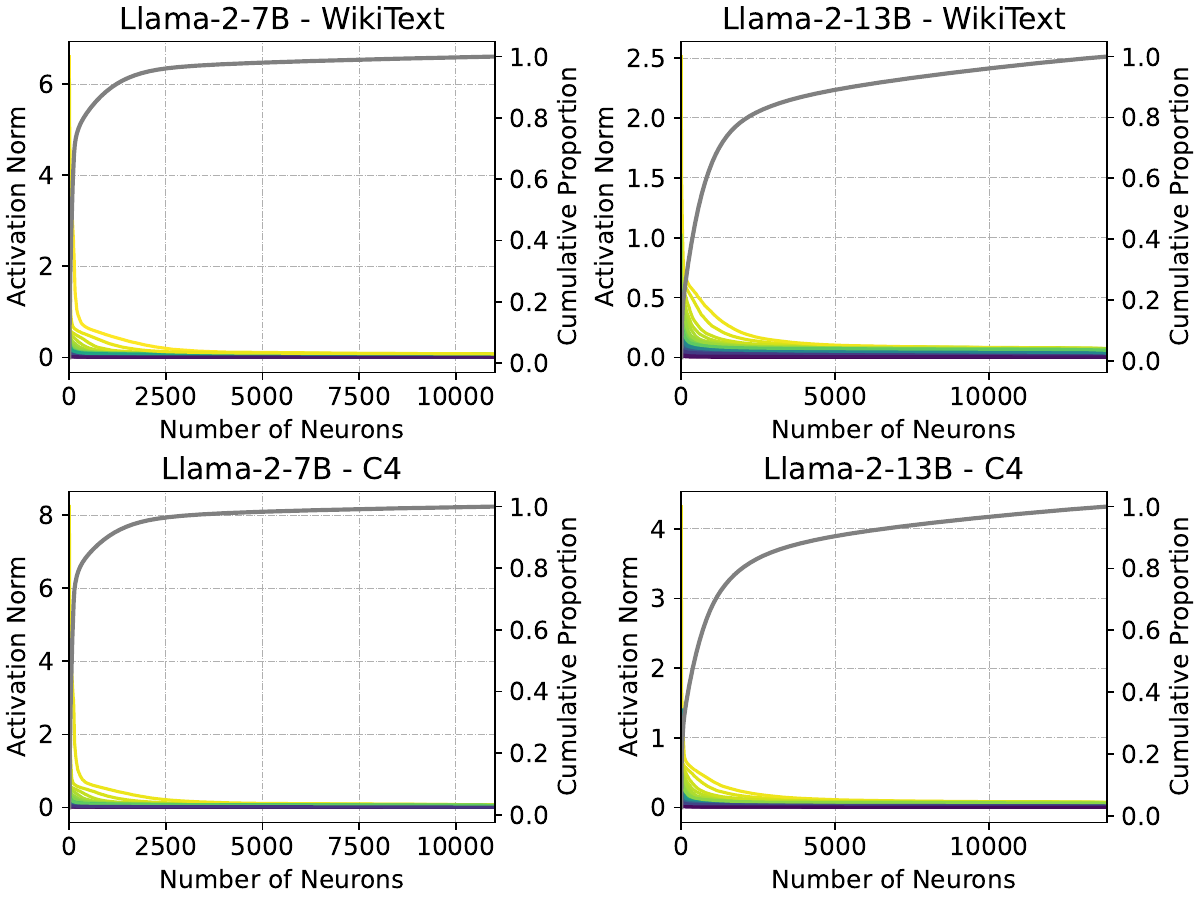} 
\caption{Accumulation of $\Vert X W \Vert_F^2$ and distribution of 
$\Vert X W \Vert_F$
across neurons in different layers of LLaMA-2-7B and LLaMA-2-13B on WikiText2 and c4 datasets, sorted from largest to smallest, highlighting the soft activation sparsity phenomenon.}
\label{fig:fig_hot_cold}
\end{figure}

\subsection{Soft Activation Sparsity in Modern LLMs}

Activation sparsity exists in neural networks with ReLU as its activation function, where the proportion of non-zero values in the outputs of ReLU activation functions is remarkably low. It also exists in many earlier LLMs that adopt ReLU as its activation in the FFN, such as OPT~\citep{opt} and GPT~\citep{gpt}. Activation sparsity has been extensively studied to improve inference quality and efficiency ~\citep{dejavu, LearnBeEfficient2024}. However, as for modern LLMs, we can no longer exploit this feature since soft activation functions, \eg SiLU and GeLU, are widely used to replace ReLU, where neurons still remain activated when inputs are below zero.

To identify if there is any activation pattern in modern LLMs that is similar to activation sparsity, we examine the distribution of activation norms in LLaMA-2-7B/13B~\cite{llama_2} on WikiText2~\cite{wikitext} and C4~\cite{c4}. 
As depicted in Figure~\ref{fig:fig_hot_cold}, activation norms of a certain group of neurons occupy most of the total and the rest are nearly round to 0. It indicates that long-tail distribution exists in the activation norms of the FFN. Intuitively, the importance of different neurons can be denoted by their corresponding activation magnitude. To verify the presumption, we thus investigate how much neurons contribute to the model performance by eliminating the highest or lowest neurons. The model performance is evaluated through computing perplexity on WikiText2.

We summarize the evaluation results in Table~\ref{tab:neuron_removal}. It shows neurons that have the highest activation norms contribute the most to the model performance, and removing them can severely deteriorate model performance. As for their counterpart, removing them does not bring such significant performance degradation as much as removing the highest ones. \textit{Therefore, we conclude that soft activation sparsity exists in the FFN of modern LLMs, where activation norms of a certain small group of neurons occupy most of the total, and removing the corresponding neurons can lead drastic performance drop.}

\begin{table}[h]
\centering
\begin{tabular}{c|c|c|cccc}
\toprule 
    \multirow{2}{*}{\shortstack{LLaMA- \\ 2-13B}} & \multirow{2}{*}{original} & \multicolumn{1}{c}{PN} & \multicolumn{3}{c}{MN} \\
    \cmidrule{3-6}
    & & 1\% & 10\% & 30\% & 50\% \\
    \hline
    perplexity ($\downarrow$)  & 4.57 & 9665.4 & 4.83 & 6.58 & 17.03 \\
\bottomrule
\end{tabular}
\caption{Impact of neuron pruning on LLaMA-2-13B model perplexity, highlighting the sensitivity to the loss of high-norm ``Prime Neurons" (PN) and the resilience following the removal of low-norm ``Marginal Neurons" (MN).}
\label{tab:neuron_removal}
\end{table}

\subsection{Soft Activation Sparsity Driven Decomposition}

To capture data distribution of inputs and outputs, model decomposition in our proposed method generally follows instructions described by~\citet{svd_llm}. Initially, we prepare calibration data and collect input $X$ of each layer, then perform Cholesky decomposition on $XX^T$ to get the scaling matrix $S$. In the end, $WS^{-1}$ is being decomposed with SVD: $WS^{-1} = U \Sigma V$. Additionally, motivated by the existence of soft activation sparsity in modern LLMs, we improve the model decomposition quality by refining the FFN decomposition with exploitation of soft activation sparsity.

To refine the FFN decomposition, the neurons are first sorted according to their activation norms in descendent order and then grouped into two clusters. 
Those that tend to produce higher activation norms are coined as ``prime neurons'' (PN), and the rest are coined as ``marginal neurons'' (MN). 
The grouping criterion is controlled by a hyperparameter $\gamma$, \ie the ratio of PN. 
We can utilize the accumulated squared L2 Norm to identify $\gamma$. For instance, in LLaMA-2-13B, the top 15\% of neurons occupy 95\% of the total. Then $\gamma$ can be set to 0.15.
The computing of the FFN can be rewritten as Equation~\eqref{eq:ffn}.
\begin{equation}
\begin{aligned}
FFN(X) &= \sigma(X W^{in}) \times W^{out} \\ 
&= \sigma(X W_{\alpha}^{in}) W_{\alpha}^{out} + \sigma(X W_{\beta}^{in}) W_{\beta}^{out}
\end{aligned}
\label{eq:ffn}
\end{equation}
where $W_\alpha$ denotes the subset of the weight matrix corresponding to PN, $W_\beta$ denotes the rest of MN, and $X$ is the input.

As removing important neurons could lead to drastic performance degradation, we thus retain these neurons and only decompose the less important ones,  i.e., $W_{\beta}$. Moreover, to capture data distribution in inputs and outputs, we partition the scaling matrix $S$ into $S_{\alpha}$ and $S_{\beta}$ according to the partition of neurons, and then employ SVD to decompose $W_{\beta}$, \ie$U_{\beta} \Sigma_{\beta} V_{\beta}= W_{\beta}S_{\beta}^{-1}$. Thus, the computing of the less important neurons can be formulated as follows.
\begin{equation}
\begin{aligned}
\sigma(X W_{\beta}^{in}) W_{\beta}^{out} = \sigma(X U_{\beta}^{in} \Sigma_{\beta}^{in} V_{\beta}^{in}) U_{\beta}^{out} \Sigma_{\beta}^{out} V_{\beta}^{out}
\end{aligned}
\label{eq:ffn_decompose}
\end{equation}

The attention module also exhibits sparse property~\cite{dejavu, flap}. But it tends not to use the activation function to induce nonlinear transformations. Consequently, we employ low-rank decomposition to compress the entire set of weight matrices within the attention module.

\begin{table*}[t]
\centering
\begin{tabular}{l|c|c|cccccccc}
\toprule
Methods& Ratio & Average & MMLU & BoolQ& PIQA& WinoGrande& HellaSwag& ARC-e& ARC-c& OBQA\\
\hline

LLaMA-2-7B& 0\% & 0.6410 & 0.457 & 0.7777 & 0.7905 & 0.6938 & 0.7592 & 0.7449 & 0.4625 & 0.442 \\
\hline
LLM-Pruner& \multirow{6}*{20\%} & 0.5512 & 0.262 & 0.6376 & \textbf{0.7595} & 0.6338 & \textbf{0.6783} & 0.6431 & \textbf{0.3993} & 0.396 \\
FLAP&  & 0.5318 & 0.319 & 0.5394 & 0.7454 & 0.6298 & \underline{0.6474} & 0.6128 & 0.3643 & \underline{0.396} \\
SliceGPT&  & 0.4184 & 0.263 & 0.3792 & 0.6126 & 0.5983 & 0.4428 & 0.4609 & 0.2841 & 0.306 \\
Bolaco&  & \textbf{0.5733} & \textbf{0.343} & \underline{0.7201} & \underline{0.7509} & \underline{0.6561} & 0.6433 & \textbf{0.6819} & \underline{0.3748} & \textbf{0.416} \\
SVD-LLM&  & 0.4673 & 0.268 & 0.5468 & 0.6513 & 0.6243 & 0.5173 & 0.4722 & 0.2782 & 0.380\\
SoLA (Ours) &  & \underline{0.5692} & 0.341 & \textbf{0.7505} & 0.7465 & \textbf{0.6646} & 0.6392 & \underline{0.6561} & 0.3737 & 0.382 \\
\hline
LLM-Pruner& \multirow{6}*{30\%}& 0.4767 & 0.246 & 0.5324 & \textbf{0.7225} & 0.5454 & \underline{0.5696} & 0.5109 & 0.3166 & 0.370 \\
FLAP&  & 0.4893 & 0.267 & 0.5220 & 0.7029 & \underline{0.6006} & 0.5658 & 0.5518 & \underline{0.3225} & \underline{0.382} \\
SliceGPT&  & 0.3757 & 0.259 & 0.3783 & 0.5555 & 0.5446 & 0.3517 & 0.3906 & 0.2457 & 0.280 \\
Bolaco&  & \underline{0.5138} & \textbf{0.280} & \textbf{0.7008} & \underline{0.7184} & 0.5917 & 0.5361 & \underline{0.5871} & 0.3077 & \textbf{0.388}\\
SVD-LLM&  & 0.4252 & 0.255 & 0.5180 & 0.6001 & 0.5825 & 0.4185 & 0.4331 & 0.2543 & 0.340\\
SoLA (Ours)&  & \textbf{0.5157} & \underline{0.277} & \underline{0.6673} & 0.6997 & \textbf{0.6283} & \textbf{0.5711} & \textbf{0.5913} & \textbf{0.3268} & 0.364 \\

\hline\hline

LLaMA-2-13B& 0\%& 0.6756 & 0.554 & 0.8055 & 0.8041 & 0.7253 & 0.7941 & 0.7739 & 0.4915 & 0.456\\
\hline
LLM-Pruner& \multirow{6}*{20\%}& 0.5639 & 0.228 & 0.6297 & \textbf{0.7797} & 0.6077 & \textbf{0.7126} & 0.6709 & \textbf{0.4428} & \underline{0.440} \\
FLAP&  & 0.5818 & 0.412 & 0.6642 & 0.7557 & 0.6725 & 0.6919 & 0.6591 & 0.3908 & 0.408 \\
SliceGPT&  & 0.4488 & 0.310 & 0.3786 & 0.6224 & 0.6354 & 0.4730 & 0.4659 & 0.3191 & 0.386 \\
Bolaco&  & \underline{0.6138} & \underline{0.434} & \underline{0.7649} & \underline{0.7683} & 0.6590 & \underline{0.6996} & \textbf{0.7093} & \underline{0.4272} & \textbf{0.448}\\
SVD-LLM&  & 0.5574 & 0.346 & 0.7217 & 0.716 & \underline{0.6843} & 0.5991 & 0.6212 & 0.3669 & 0.404 \\
SoLA (Ours)&  & \textbf{0.6142} & \textbf{0.461} & \textbf{0.7951} & 0.7557 & \textbf{0.6977} & 0.6735 & \underline{0.6915} & 0.407 & 0.432 \\
\hline
LLM-Pruner& \multirow{6}*{30\%}& 0.5090 & 0.229 & 0.6211 & 0.7318 & 0.5793 & 0.6089 & 0.5471 & 0.3404 & \textbf{0.414} \\
FLAP&  & 0.5429 & 0.332 & 0.6437 & 0.7242 & 0.6393 & \textbf{0.6244} & 0.6145 & \underline{0.3729} & 0.392 \\
SliceGPT&  & 0.3954 & 0.271 & 0.3783 & 0.5675 & 0.5770 & 0.3827 & 0.4087 & 0.2619 & 0.316 \\
Bolaco&  & \underline{0.5608} & \underline{0.343} & \underline{0.7504} & \underline{0.7246} & \underline{0.6446} & 0.5773 & \textbf{0.6560} & \textbf{0.3919} & 0.398\\
SVD-LLM&  & 0.4854 & 0.286 & 0.6401 & 0.6556 & 0.6393 & 0.4800 & 0.5059 & 0.3003 & 0.376\\
SoLA (Ours)&  & \textbf{0.5756} & \textbf{0.394} & \textbf{0.7713} & \textbf{0.7263} & \textbf{0.6740} & \underline{0.6138} & \underline{0.6557} & 0.3677 & \underline{0.402} \\

\hline\hline

LLaMA-2-70B& 0\%& 0.7294 & 0.688 & 0.8388 & 0.8275 & 0.7782 & 0.838 & 0.8072 & 0.5717 & 0.486\\
\hline
FLAP& \multirow{4}*{20\%} & 0.5003 & 0.259 & 0.6226 & 0.7231 & 0.6409 & 0.5594 & 0.5105 & 0.3191 & 0.368 \\
SliceGPT&  & 0.5572 & 0.483 & 0.4394 & 0.6801 & 0.7214 & 0.5716 & 0.6864 & 0.4394 & \underline{0.436}\\
SVD-LLM&  & \underline{0.6275} & \underline{0.521} & \underline{0.7453} & \underline{0.7448} & \underline{0.7261} & \underline{0.6841} & 0.7193 & \underline{0.4693} & 0.410\\
SoLA (Ours)&  & \textbf{0.6892} & \textbf{0.624} & \textbf{0.7483} & \textbf{0.7911} & \textbf{0.7656} & \textbf{0.7751} & \textbf{0.7963} & \textbf{0.5452} & \textbf{0.468}\\
\hline
FLAP& \multirow{4}*{30\%} & 0.4962 & 0.264 & 0.6526 & \underline{0.6959} & 0.6480 & 0.5561 & 0.4891 & 0.3055 & 0.358 \\
SliceGPT&  & 0.4635 & 0.326 & 0.3783 & 0.6235 & 0.6701 & 0.4491 & 0.5404 & 0.3285 & 0.392\\
SVD-LLM&  & \underline{0.6091} & \underline{0.445} & \underline{0.6869} & 0.6948 & \underline{0.6914} & \underline{0.5992} & \underline{0.6974} & \underline{0.4488} & \underline{0.410}\\
SoLA (Ours)&  & \textbf{0.6625} & \textbf{0.570} & \textbf{0.7251} & \textbf{0.7791} & \textbf{0.7561} & \textbf{0.7197} & \textbf{0.7757} & \textbf{0.5222} & \textbf{0.452}\\

\bottomrule
\end{tabular}
\caption{Downstream task accuracy of the  compressed LLaMA-2-7B/13B/70B models. 
\textbf{Bold} denotes the best result at the same compression ratio, while \underline{underline} indicates the second best result. }
\label{tab:zeroshot}
\end{table*}

\subsection{Component-wise Truncation Position}

Extensive studies~\citep{bolaco, WeLore} have demonstrated that there are inherent differences among components. Components of different types thus have different sensitivities to decomposition. Therefore, it is necessary for component-wise truncation position selection rather than simply adopting a uniform truncation position setting.

\begin{theorem}
\label{thm:svd-llm}
Given an input $X$, a weight matrix $W$ and its singular value decomposition results from $U \Sigma V^T=W$. Let $S$ be the Cholesky decomposition of $XX_T$. The compression loss of truncating the smallest singular values is $L^2=\Vert \sum_{i=m+1}^{r} \sigma_i u_i v_{i}^{T} S^{-1} X \Vert_{F}^{2}=\sum_{i=m+1}^{r}(\sigma_i)^2$ and such truncating leads to the lowest loss.
\end{theorem}

To this end, we devise an adaptive component-wise allocation strategy to handle the task of truncation position determination. Our method is based on the closed-form solution of the reconstruction error given by Theorem~\ref{thm:svd-llm}~\citep{svd_llm}. We define the performance score of compressed layers as Equation~\eqref{eq:perf_score} below.
\begin{equation}
  f(r) = \frac{\sum_{i=0}^r \sigma_i^2}{\sum \sigma^2}
  \label{eq:perf_score}
\end{equation}
where $\sigma$ denotes singular values of $WS^{-1}$ and $r$ is the truncation position.

Concerning a memory budget (\ie compression rate), we can formulate the following optimization problem:
\begin{equation}
\begin{aligned}
\mathop{\text{argmax}}_{r}\sum & f(r_c) \\
    s.t. \sum{g(r_c)} &\leq \mathcal{B} \\
\end{aligned}
\end{equation}
where $r_c$ denotes the truncation position of component $c$, $g(r_c)$ denotes the memory occupation of component $c$ under its truncation position $r_c$, and $\mathcal{B}$ is the memory budget.

This optimization problem is an integer programming problem and performing an exhaustive search in an enormous solution space is infeasible. Therefore, 
We employ an adaptive heuristic greedy search algorithm, which dynamically selects the desired truncation position for each component as directed by the performance function, thereby obtaining a sub-optimal solution within an acceptable searching time.
To leverage NVIDIA hardware acceleration, the $r$ is set to multiples of 16~\cite{features_low_rank}.

\section{Experiments}

Here, we investigate our proposed SoLA across various benchmarks using different LLM series at three scales. Furthermore, we present in-depth studies of SoLA.

\subsection{Experimental Settings}

We evaluate SoLA over different series and scales of LLMs: LLaMA-2 7B, 13B, and 70B, as well as Mistral-7B-v0.1.
The language modeling capability is evaluated on the WikiText2~\cite{wikitext} test set.
We use Language Model Evaluation Harness~\cite{gao2021framework} to assess zero-shot common sense reasoning performance. 
Moreover, the 5-shot Massive Multitask Language Understanding (MMLU) accuracy~\cite{mmlu} is used for the evaluation. 
We compare SoLA with the state-of-the-art structured pruning and low-rank decomposition methods discussed in related works, including LLM-Pruner, FLAP, SliceGPT, Bolaco, and SVD-LLM.

\subsection{Overall Performance}

We evaluate the performance of compressed models by each compression method at different compression ratios ranging from 20\% to 50\%.
The perplexity scores for language modeling are shown in Figure \ref{fig:ppl} and Table \ref{tab:ppl_results}, the zero-shot common sense reasoning results and the 5-shot MMLU accuracy of LLaMA-2 series are in Table \ref{tab:zeroshot}.
The results of Mistral-7B are listed in Appendix A Table 1. LLM-Pruner and Bolaco are currently not suitable for the GQA architecture such as LLaMA-2-70B and Mistral-7B.

\subsubsection{Language Modeling}
\begin{figure}[t]
\centering
\includegraphics[width=0.97\columnwidth]{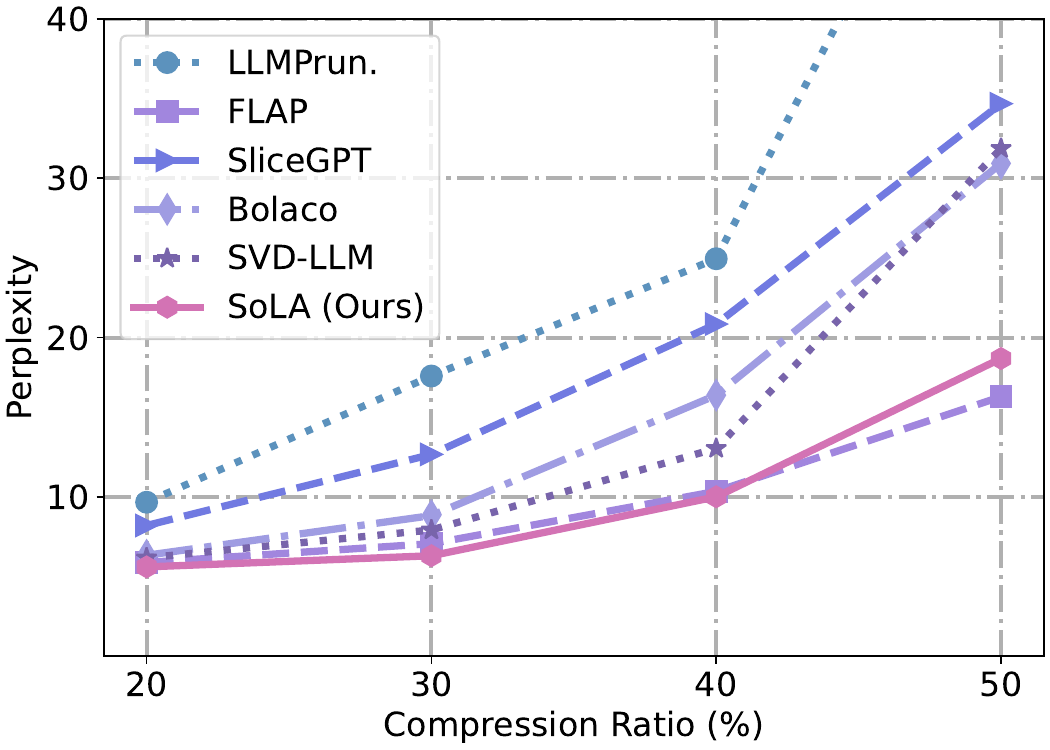} 
\caption{Perplexity of WikiText2 among different methods on LLaMA-2-13B.}
\label{fig:ppl}
\end{figure}

As shown in Figure \ref{fig:ppl}, SoLA performs remarkable perplexity. As the compression ratio increases, perplexity grows slowly, indicating a better capability to maintain model generation capability. 
In contrast, the quality of baseline methods such as LLM-Pruner sharply declines as the compression ratio increases, particularly when the pruning ratio exceeds 40\%, requiring fine-tuning to achieve acceptable performance.
SoLA narrows the performance gap between the compressed model and the original model in almost all configurations, and only FLAP slightly surpasses SoLA at LLaMA-2-13B compression rate above 40\%, demonstrating the strong competitiveness of SoLA.

\subsubsection{Downstream Tasks Performance}

For zero-shot and five-shot downstream scenarios, excluding the 20\% compression ratio in LLaMA-2-7B, SoLA consistently demonstrates superior performance over all baseline methods, achieving a 3\% to 10\% improvement in average accuracy compared to baseline methods.

\begin{table}[t]
\centering
\small
\begin{tabular}{c|c|cccc}
\toprule 
    \multirow{2}{*}{Method} & 
    \multirow{2}{*}{Ratio} & \multicolumn{3}{c}{LLaMA-2} & \multicolumn{1}{c}{Mistral} \\
    \cmidrule{3-6}
    & & 7B & 13B & 70B & 7B \\
    \hline
    Dense  & 0$\%$ & 5.11 & 4.57 & 3.12 & 4.92 \\
    
    \hline
    LLM-Pruner & \multirow{7}{*}{20$\%$}  & 10.55 & 9.67 & - & - \\ 
    FLAP &  & 6.76 & 5.90 & 8.76 & 7.11 \\
    SliceGPT &  & 9.70 & 8.21 & 5.76 & 8.23 \\
    Bolaco &  & 7.31 & 6.34 & - & - \\
    SVD-LLM &  & 8.07 & 6.18 & 5.96 & 7.26 \\
    SoLA (Ours) &  & \textbf{6.52} & \textbf{5.61} & \textbf{4.06} & \textbf{6.06} \\
    \hline
    
    LLM-Pruner & \multirow{7}{*}{30$\%$} & 18.25 & 17.59 & - & - \\ 
    FLAP &  & 8.91 & 7.08 & 10.80 & 13.10 \\
    SliceGPT &  & 15.42 & 12.68 & 8.09 & 14.69 \\
    Bolaco &  & 12.19 & 8.83 & - & - \\
    SVD-LLM &  & 11.40 & 7.93 & 6.95 & 12.32 \\
    SoLA (Ours) &  & \textbf{7.81} & \textbf{6.31} & \textbf{4.44} & \textbf{7.38} \\
\bottomrule
\end{tabular}
\caption{WikiText2 validation perplexity of pruning methods for LLaMA-2 model series and Mistral-7B-v0.1.}
\label{tab:ppl_results}
\end{table}

\subsection{In-Depth Analysis}

We present extensive studies on two fundamental components of SoLA:
soft activation sparsity driven decomposition and component-wise truncation position.
Furthermore, we evaluate the robustness of SoLA to calibration samples.
We pose the following research questions:  
Q1: What is the significance of ``Prime Neurons" in balancing the trade-off between accuracy and efficiency in compressed LLMs, and how should the ratio of “Prime Neurons” be determined?
Q2: What effect does the adaptive component-wise rank allocation strategy have?
Q3: How does the sensitivity of SoLA vary with the type and number of the calibration dataset?

\begin{figure}[t]
    \centering
    \includegraphics[width=0.99\linewidth]{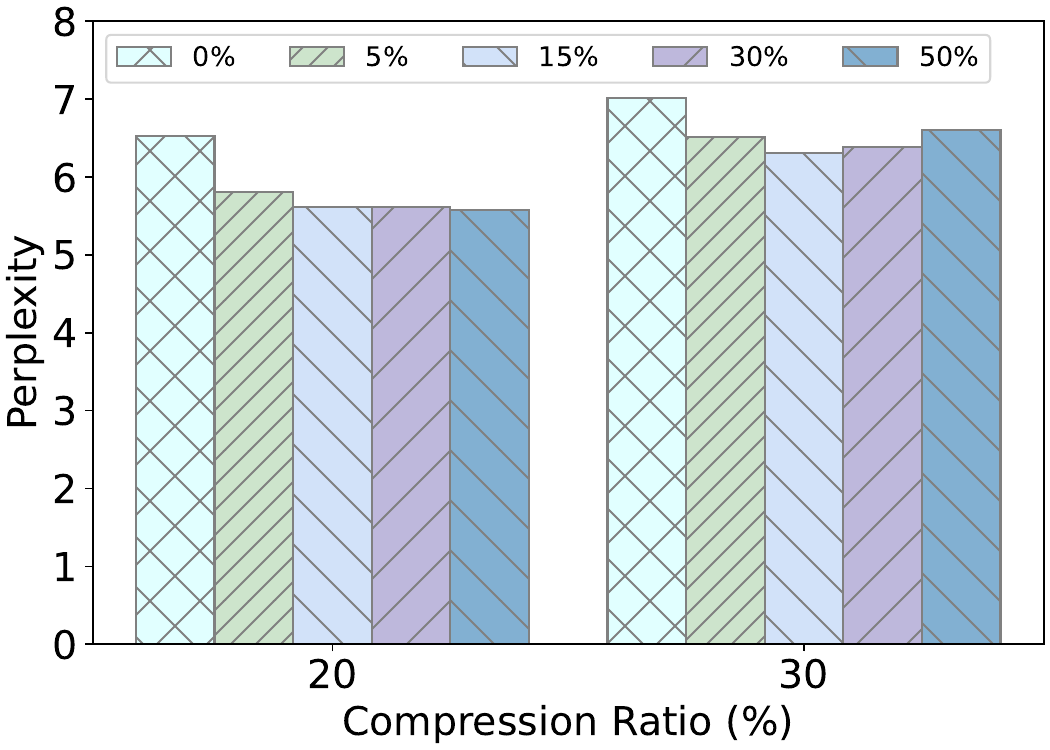}
    \caption{
    The impact of ``Prime Neurons" ratios on LLaMA-2-13B perplexity under 20\% and 30\% compression ratios.
    }
    \label{fig:hot_ratio}
\end{figure}

\subsubsection{Impact of Prime Neurons}

We first validate the importance of “Prime Neurons” (PN), setting the portion of PN to 0\%. 
Furthermore, we explore the impact of the portion of PN. 
We define four ratios for PN: 5\%, 15\%, 30\%, and 50\%, and then compare the perplexity at the same pruning ratio (20\% and 30\%). 
Experiments are conducted on LLaMA-2-13B and detailed results are shown in Figure~\ref{fig:hot_ratio}. 

It can be observed that maintaining only 5\% of PN, 
can lead to a significant improvement in perplexity (5.8 vs. 6.5 under 20\% compression ratio). 
This finding validates the conclusion drawn in earlier: a small proportion of large output norm neurons in the FFN significantly contribute to performance, while the remaining neurons can be compressed.
The 15\% configuration serves as the default configuration in the experimental section.

\subsubsection{Contribution of Adaptive Rank Allocation}

\begin{table}[ht]
\centering
\small
\begin{tabular}{c|c|cc|cc}
\toprule 
    \multirow{2}{*}{Model} & \multirow{2}{*}{Ratio} & \multicolumn{2}{c}{Perplexity ($\downarrow$)} & \multicolumn{2}{c}{Avg. Acc. ($\uparrow$)} \\
    \cmidrule{3-6}
    & & Unif. & Adap. & Unif. & Adap. \\
    \midrule
    LLaMA-2-7B & \multirow{3}{*}{20$\%$}  & 8.07 & \textbf{7.18} & 0.467 & \textbf{0.541} \\
    LLaMA-2-13B & & \textbf{6.18} & 6.52 & 0.557 & \textbf{0.564} \\
    Mistral-7B  & & 7.26 & \textbf{6.68} & 0.528 & \textbf{0.578} \\
    \midrule
    LLaMA-2-7B & \multirow{3}{*}{30$\%$} & 11.40 & \textbf{9.32} & 0.425 & \textbf{0.492} \\
    LLaMA-2-13B &  & 7.93 & \textbf{7.02} & 0.485 & \textbf{0.541} \\
    Mistral-7B &  & 12.32 & \textbf{10.09} & 0.432 & \textbf{0.491} \\
\bottomrule
\end{tabular}
\caption{Comparison of perplexity and average accuracy of downstream tasks between uniform and adaptive strategy.} 
\label{tab:ablation_rank}
\end{table}

The uniform rank allocation method assigns low-rank dimensions to all components based on the target compression rate, \eg $gate/up/down$ projections in the FFN use the same rank $r = target\_rate \times (m \times n)/(m+n)$. In contrast, our adaptive component-wise rank allocation strategy considers the compression sensitivity of each component.
Table~\ref{tab:ablation_rank} demonstrates that our adaptive strategy improves perplexity by 8\%-18\% and downstream task average accuracy up to 14\%.

\subsubsection{Robustness to Calibration Dataset}

\begin{figure}[ht]
\centering
\includegraphics[width=0.99\columnwidth]{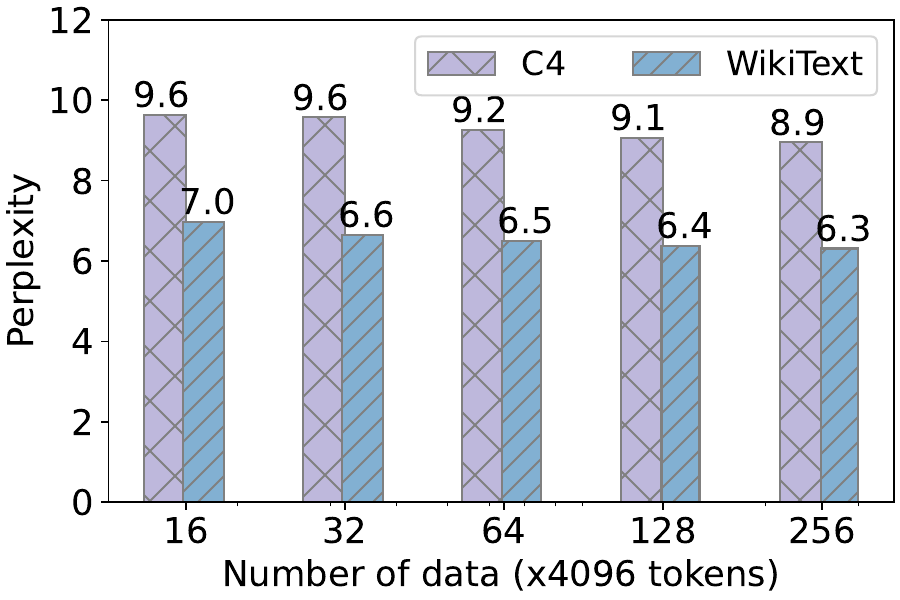} 
\caption{Perplexity of LLaMA-2-13B under 30\% compression ratio using calibration data with different numbers (32, 64, 128, 256) and types (WikiText2 and C4).}
\label{fig:ablation_data}
\end{figure}

Finally, we examine the effect of calibration data, which captures activation patterns and influences low-rank decomposition. 
The analysis is conducted by varying the quantity and category of calibration data.
Figure \ref{fig:ablation_data} illustrates the perplexity scores on the WikiText2 test dataset resulting from the compression of LLaMA-13B.
The variations in performance due to different quantities do not exceed 10\% and perplexity degradation caused by types of calibration data is also limited, indicating SoLA is robust to the calibration data.

\subsection{Inference Efficiency}

Each LLaMA-2 block contains a feed-forward module with $gate/up/down$ operation and an attention module with $q/k/v/o$ operation.
We choose a sequence length of 2048, replicating the size of the matrix-matrix multiplications in three different-sized LLaMA-2 models.
We take the median runtime over $10^3$ attempts on RTX4090.
Table \ref{tab:ops_time} shows the total time taken in $ms$ and the corresponding speedup, each matrix multiplication cost is shown in Appendix A Table 2. 
At a 20\% pruning ratio, SoLA accelerates the matrix multiplication speed by 1.4$\times$, at a 30\% pruning ratio, it accelerates the matrix multiplication speed by 1.7$\times$.
The acceleration is achieved by replacing large weight matrices with decomposed smaller matrices and leverages existing hardware capabilities (\ie dense kernels).

\begin{table}[t]
\centering
\small
\begin{tabular}{c|ccc}
\toprule 
    % Compression \\ 
    \multirow{2}{*}{\shortstack{Ratio}} & \multicolumn{3}{c}{\shortstack{Total Time of Operation (speedup)}} \\
    & 7B & 13B & 70B \\
    \midrule
    0\%  & 20.04 & 31.64 & 96.58 \\
    20\%& 16.39 (1.22$\times$) & 21.92 (1.44$\times$) & 65.76 (1.47$\times$) \\
    30\%& 13.04 (1.54$\times$) & 17.87 (1.77$\times$) & 57.04 (1.69$\times$) \\
\bottomrule
\end{tabular}
\caption{Operation cost of LLaMA-2 series.}
\label{tab:ops_time}
\end{table}

\section{Limitations and Future Work}

Our proposed approach can be easily integrated with existing methods for measuring layer significance~\cite{owl}, achieving layer-wise compression; (ii) our work holds the potential to be integrated into inference frameworks to facilitate acceleration of end-to-end inference time.

\section{Conclusion}

In this work, we propose SoLA, a novel training-free compression method leveraging \textbf{So}ft activation sparsity and \textbf{L}ow-r\textbf{A}nk decomposition.
SoLA is built on our analysis of the activation pattern in the feed-forward network of modern LLMs and achieves fine-grained low-rank compression, which preserves a minority of significant components and compresses the majority through Singular Value Decomposition (SVD).
To alleviate the decomposition loss, we propose an adaptive component-wise low-rank allocation strategy by formulating it as an integer programming problem. Through the allocation of appropriate ranks to different types of weight matrices, our strategy enhances model quality after compression. 
Our comprehensive experiments conducted on the LLaMA-2 series and Mistral reveal that SoLA, without post-training, outperforms current state-of-the-art methods in language modeling and downstream tasks.

\appendix

\begin{table*}[t]
\centering
\begin{tabular}{l|c|c|cccccccc}
\toprule
Methods& Ratio & Average & MMLU & BoolQ& PIQA& WinoGrande& HellaSwag& ARC-e& ARC-c& OBQA\\
\hline

Mistral-7B& 0\% & 0.701 & 0.625 & 0.8398 & 0.8205 & 0.7395 & 0.8102 & 0.7955 & 0.5392 & 0.44\\
\hline
FLAP& \multirow{4}*{20\%} & 0.500 & 0.259 & 0.6226 & 0.7231 & 0.6409 & 0.5594 & 0.5105 & 0.3191 & \underline{0.368}\\
SliceGPT&  & 0.427 & 0.286 & 0.3786 & 0.6066 & 0.5943 & 0.4510 & 0.4815 & 0.3003 & 0.320\\
SVD-LLM&  & \underline{0.578} & \underline{0.418} & \textbf{0.6829} & \underline{0.7339} & \underline{0.6843} & \underline{0.6175} & \textbf{0.7134} & \textbf{0.4053} & 0.366\\
SoLA (Ours)&  & \textbf{0.581} & \textbf{0.442} & \underline{0.6609} & \textbf{0.7367} & \textbf{0.6875} & \textbf{0.6332} & \underline{0.6999} & \underline{0.3976} & \textbf{0.392}\\
\hline
FLAP& \multirow{4}*{30\%} & \underline{0.496} & 0.264 & \textbf{0.6526} & \textbf{0.6959} & \textbf{0.6480} & \textbf{0.5561} & 0.4891 & 0.3055 & \underline{0.358}\\
SliceGPT&  & 0.358 & 0.25 & 0.3783 & 0.5441 & 0.5162 & 0.3254 & 0.3502 & 0.2295 & 0.268\\
SVD-LLM&  & 0.491 & \underline{0.282} & \underline{0.6462} & 0.6491 & 0.6417 & 0.4736 & \underline{0.5825} & \underline{0.3072} & 0.342\\
SoLA (Ours)&  & \textbf{0.517} & \textbf{0.338} & 0.6257 & \underline{0.6839} & \underline{0.6448} & \underline{0.5300} & \textbf{0.6090} & \textbf{0.3276} & \textbf{0.376}\\

\bottomrule
\end{tabular}
\caption{Downstream task accuracy of the compressed Mistral-7B models. 
\textbf{Bold} denotes the best result at the same compression ratio, while \underline{underline} indicates the second best result. }
\label{tab:zeroshot_mistral}
\end{table*}

\begin{table*}[t]
\centering
\small
\begin{tabular}{c|c|ccccccc}
\toprule 
    \multirow{2}{*}{Model} & \multirow{2}{*}{\shortstack{Compression \\ Ratio}} & \multicolumn{7}{c}{Operations($ms$)} \\
    & & Gate & Up & Down & Q & K & O & Total (speedup) \\
    \midrule
    \multirow{3}{*}{LLaMA-2-7B} & Dense  & 4.94 & 4.79 & 5.09 & 1.75 & 1.74 & 1.74 & 20.04 \\
    & 20\%& 2.92 & 3.52 & 6.66 & 0.87 & 0.69 & 1.74 & 16.39 (1.22$\times$) \\
    & 30\%& 2.92 & 3.00 & 4.48 & 0.69 & 0.69 & 1.27 & 13.04 (1.54$\times$) \\
    \midrule
    \multirow{3}{*}{LLaMA-2-13B} & Dense  & 7.14 & 7.12 & 8.11 & 3.07 & 3.09 & 3.12 & 31.64 \\
    & 20\%& 5.61 & 5.08 & 5.44 & 1.34 & 1.34 & 3.12 & 21.92 (1.44$\times$) \\
    & 30\%& 5.19 & 4.53 & 4.77 & 0.93 & 0.68 & 1.77 & 17.87 (1.77$\times$) \\
    \midrule
    \multirow{3}{*}{LLaMA-2-70B} & Dense  & 23.89 & 23.85 & 26.69  & 7.33 & 7.36 & 7.47 & 96.58 \\
    & 20\%& 12.74 & 16.00 & 18.99 & 3.20 & 7.36 & 7.47 & 65.76 (1.47$\times$) \\
    & 30\%& 11.45 & 13.70 & 18.99 & 1.45 & 7.36 & 4.08 & 57.04 (1.69$\times$) \\
\bottomrule
\end{tabular}
\caption{Results of timing the matrix multiplications in each component of LLaMA-2 series.}
\label{tab:ops_time_all}
\end{table*}

\section{A \quad Additional Experiments}

\subsection{Compression on Mistral-7B}

We evaluate the zero-shot common sense reasoning performance and 5-shot Massive Multitask Language Understanding (MMLU) accuracy. Table \ref{tab:zeroshot_mistral} presents the detailed results of Mistral-7B using different compression methods. Our approach demonstrates performance improvements over state-of-the-art methods.

\subsection{Inference Efficiency of Components}

We chose a sequence length of 2048, replicating the size of the matrix-matrix multiplications in LLaMA-2 series. We take the median runtime over $10^3$ attempts on RTX4090. Table \ref{tab:ops_time_all} shows the time taken in $ms$ to run matrix multiplication of each component in the model.

\section{B \quad Implementation Details} \label{app:imp_details}

Our method can be integrated into existing low-rank decomposition techniques. We use SVD-LLM~\cite{svd_llm} for model decomposition. Following calibration setups in previous works, our calibration setup involved randomly selecting 256 samples from the training sets of WikiText2 and C4 as calibration data, with each sample having a sequence length of 4,096.

Prior work~\cite{bolaco} has demonstrated that compressing the weight matrix of the $v$ projection in the attention module leads to significant performance degradation, hence we exclude the $v$ projection from compression. 
In the case of LLaMA-2-7B and LLaMA-2-13B, the $o$ projection remains uncompressed at a rate of 20\%.
For LLaMA-2-70B and Mistral-7B models utilizing group query attention, both the $k$ and $v$ projections are not subjected to compression. 

The initial and terminal layers of LLMs play an important role in maintaining model performance, such as the shallower layers performing feature extraction~\cite{shallow_layer}, which is why some methods do not compress these layers. For instance, LLM-Pruner leaves the first four and the final layers unaltered. Similarly, our method also avoids modifying the first and last two layers.

Due to the introduction of an additional Q matrix, the actual number of parameters in the SliceGPT model is greater than the pruned number of parameters set. To ensure a fair comparison, our compression ratio refers to the memory size of the compressed model divided by the memory size of the original model.
The FLAP should modify its masking implementation when it is used in GQA architecture models.

\section{Acknowledgments}

This work is supported by the Guangzhou Industrial Information and Intelligent Key Laboratory Project (No. 2024A03J0628), the Guangzhou Science and Technology Development Projects (No. 2023A03J0143 and No. 2024A04J4458), and the NSFC Project (No. 62306256).

\bibliography{SoLA}

\end{document}